\documentclass[runningheads]{llncs}
\usepackage[T1]{fontenc}
\usepackage{graphicx,verbatim}
\usepackage{amsmath,amssymb,amsfonts}
\usepackage{algorithmic}
\usepackage[noadjust]{cite}
\usepackage{multirow}
\usepackage{tabularray}
\usepackage{booktabs}
\usepackage{pifont}
\usepackage{color,soul}

\usepackage{booktabs}
\usepackage{array}
\usepackage{xspace}
\usepackage{hyperref}
\hypersetup{
    colorlinks=true,
    urlcolor=cyan,
    }
\usepackage[capitalise]{cleveref}

\usepackage{color}

\newcommand{\cmark}{\ding{51}}
\newcommand{\xmark}{\ding{55}}
\newcommand*{\eg}{\emph{e.g.}\xspace}
\newcommand*{\ie}{\emph{i.e.}\xspace}
\newcommand*{\proposed}{M\&N\xspace}

\begin{document}
%
\title{Semi-Supervised 3D Medical Segmentation from 2D Natural Images Pretrained Model}

%
%
%
\author{Pak-Hei Yeung\inst{1,2} \and
Jayroop Ramesh\inst{2} \and
Pengfei Lyu\inst{3} \and
Ana Namburete\inst{2} \and
Jagath Rajapakse\inst{1}}
\authorrunning{PH. Yeung et al.}
\titlerunning{\proposed}
%
\institute{College of Computing and Data Science, Nanyang Technological University, Singapore\\
\email{pakhei.yeung@ntu.edu.sg} \and
Oxford Machine Learning in NeuroImaging Lab, University of Oxford, United Kingdom \and
Faculty of Robot Science and Engineering,
Northeastern University, China}


\maketitle              

\begin{center}
\href{https://pakheiyeung.github.io/M-N_wp/}{Project Page}
\end{center}

\begin{abstract}

This paper explores the transfer of knowledge from general vision models pretrained on 2D natural images to improve 3D medical image segmentation. 
We focus on the semi-supervised setting, 
where only a few labeled 3D medical images are available, 
along with a large set of unlabeled images.
To tackle this,
we propose a model-agnostic framework
that progressively distills knowledge from a 2D pretrained model to a 3D segmentation model trained from scratch.
Our approach, \textbf{\proposed}, involves iterative co-training of the two models using pseudo-masks generated by each other,
along with our proposed learning rate guided sampling that adaptively adjusts 
the proportion of labeled and unlabeled data in each training batch to align with the models' prediction accuracy and stability,
minimizing the adverse effect caused by inaccurate pseudo-masks.
Extensive experiments on multiple publicly available datasets
demonstrate that \proposed achieves state-of-the-art performance, outperforming thirteen existing semi-supervised segmentation approaches under all different settings.
Importantly, ablation studies show that 
\proposed remains model-agnostic, allowing seamless integration with different architectures.
This ensures its adaptability as more advanced models emerge.
The code is available at \url{https://github.com/pakheiyeung/M-N}.


\end{abstract}
%
%
%
\section{Introduction}
\label{sec:intro}
The advent of deep learning has significantly boosted the performance of 3D medical image segmentation, unarguably one of the most important tasks in medical image analysis.
However, training a deep learning model from scratch typically requires a large amount of labeled data, which can be a major bottleneck in the medical domain \cite{sli2vol}. 
In contrast, labeled data is abundant in other domains, such as 2D natural images, where massive datasets \cite{ade20k,imagenet} have been curated and utilized to train powerful vision models \cite{segformer}. 
These pretrained models have demonstrated remarkable capabilities in various computer vision tasks. 
Motivated by the success of these models, 
this paper explores the possibility of leveraging their knowledge to facilitate the 3D medical image segmentation, particularly in scenarios where only a few manual labels are available. 

Specifically, we focus on the task of semi-supervised 3D medical image segmentation, 
where only a few labeled 3D medical images are available, 
accompanied by a large set of unlabeled images. 
Recent advances in this area have achieved remarkable performance through various strategies for utilizing unlabeled data, for example teacher-student frameworks \cite{admt,bcp}, uncertainty-driven approaches \cite{duwm,cobionet}, unsupervised domain adaptation \cite{lowbridge} and Prototype- and contrastive-learning-based frameworks \cite{graphcl,ssnet}.
Another line of research related to this paper explored bridging the gap between 2D and 3D networks for 3D medical image analysis, 
primarily through network architecture design \cite{acs,m3t,qcnn}. 
While remarkable performance has been achieved, 
most of these approaches are tailored for specific types of networks.
In contrast, this work proposes a model-agnostic framework that enables the transfer of knowledge from any 2D pretrained network to any 3D segmentation network,
aiming for a more flexible and generalizable solution.

This work is motivated by our initial findings
that show using
a model pretrained on 2D natural images substantially outperforms the same network trained from scratch for a 3D medical segmentation task.
This performance gap widens when the number of labeled training data is limited.
These findings are shown in the first 3 rows in \cref{tab:la8,tab:la4,tab:ct6}.
This suggests that pretraining on natural images acquires knowledge that is transferable to 3D medical segmentation,
particularly in low-data regimes.
Building on the success of 3D-based networks \cite{swinunetr,vnet},
which have achieved state-of-the-art medical segmentation performance when trained on large labeled datasets,
we pose a fundamental question:
can we leverage the knowledge from a pretrained 2D model to improve the performance of a 3D segmentation model, 
even when training on limited labeled samples? 

\begin{figure*}[!t]
    \centering 
    \includegraphics[width=1\textwidth]{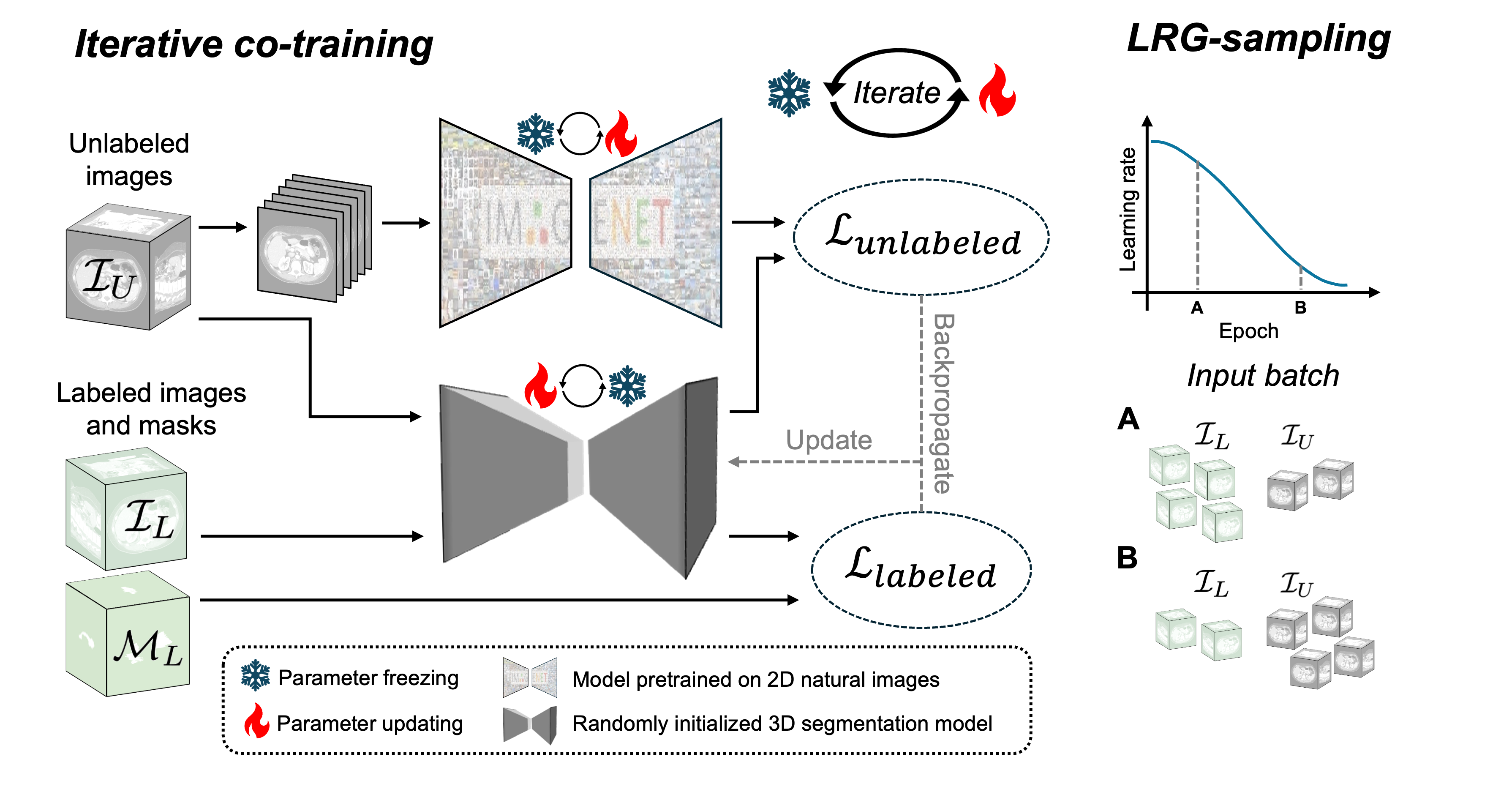}
    \caption{Pipeline of our proposed \proposed framework.
            The 2D and 3D models are iteratively
            co-trained using pseudo-masks generated by each other,
            with the unlabeled loss ($\mathcal{L}_{unlabled}$), 
            and using labeled images and masks with the labeled loss ($\mathcal{L}_{labled}$).
            This iterative process alternates between odd and even epochs.
            LRG-sampling dynamically adjusts the proportion of labeled and unlabeled data in each batch based on the current learning rate,
            optimizing the utilization of available training data.
    }
\label{fig:pipeline}
\end{figure*}

To address this question,
we propose \proposed, 
a model-agnostic framework that distills knowledge from a vision model, pretrained on 2D natural images, 
to a 3D model trained from scratch for semi-supervised medical segmentation. 
Our work makes the following contributions:
\emph{firstly,} 
we propose an iterative co-training strategy,
where the 2D and 3D models are trained using pseudo-masks generated by each other.
To mitigate the impact of inaccurate pseudo-masks,
we further propose learning rate guided sampling,
which adaptively adjusts the proportion of labeled and unlabeled data in a batch to align with the models' prediction accuracy and stability.
As our \emph{second} contribution,
we benchmark \proposed on various publicly available datasets with different limited-data settings.
\proposed outperforms 13 existing semi-supervised segmentation approaches,
achieving state-of-the-art performance on all experiments.
\emph{Thirdly},
our ablation studies show that \proposed is agnostic to different models and architectures.
This suggests its generalizability and potential for seamless integration with advanced models to achieve even more outstanding results in the future.

\section{Method}
\label{sec:method}
We propose \proposed for semi-supervised 3D medical image segmentation.
Given a dataset of 3D medical images,
$\mathcal{I} = \{{\mathbf{I}_i}\}_{i=1}^m$, 
where each image $\mathbf{I}_i \in \mathbb{R}^{C_i \times H \times W \times D}$ has $C_i$ channels, height $H$, width $W$ and depth $D$.
We assume a subset of images, $\mathcal{I}_L$,
has corresponding labeled masks,
$\mathcal{M}_L = \{{\mathbf{M}_i}\}_{i=1}^n, \mathbf{M}_i \in \mathbb{R}^{C_c \times H \times W \times D}$,
with $C_c$ classes,
where $n\ll m$.
The remaining images, $\mathcal{I}_U$, are unlabeled.

Using $\mathcal{I}_L$, $\mathcal{M}_L$ and $\mathcal{I}_U$,
our objective is to distill knowledge from a pretrained vision model, $f(\cdot;\theta_{nat})$,
parametrized by $\theta_{nat}$ and pretrained on 2D natural images,
to a 3D segmentation model,
$g(\cdot;\theta_{med})$,
parametrized by $\theta_{med}$.

\subsection{Fine-tuning on Labeled Images}
\label{sec:finetuning}
We begin by fine-tuning the pretrained 2D model, $f(\cdot;\theta_{nat})$,
on the labeled dataset $\{{\mathbf{I}_i}, \mathbf{M}_i\}_{i=1}^n$
by extracting 2D slices along the depth dimension $D$.
Simultaneously, we train the 3D segmentation model, $g(\cdot;\theta_{med})$, from scratch on $\mathcal{I}_L$ and $\mathcal{M}_L$.
During this stage, the two models are optimized independently using the labeled loss, $\mathcal{L}_l$, defined as:
\begin{equation}
\label{eq:loss_label}
\mathcal{L}_l = w_{ce}\cdot\mathcal{L}_{ce}\left(\hat{\mathbf{M}},\mathbf{M}\right) + w_{dice}\cdot\mathcal{L}_{dice}\left(\hat{\mathbf{M}},\mathbf{M}\right),
\end{equation}
where $\hat{\mathbf{M}}$ is the predicted mask, $\mathcal{L}_{ce}$ is the cross-entropy loss, $\mathcal{L}_{dice}$ is the soft Dice loss and $w_{ce}$ and $w_{dice}$ are their respective weights. 
\\
\\
\noindent \textbf{Pretrained models.}
\proposed is model-agnostic, 
allowing a wide range of 2D vision models, $f(\cdot;\theta_{nat})$, 
pretrained with different learning objectives, to be used.
For $f(\cdot;\theta_{nat})$ with an encoder-decoder architecture,
such as \cite{segformer},
fine-tuning can be done by simply replacing the last layer to match the number of classes $C_c$.
Alternatively, 
a decoder needs to be appended and fine-tuned
for models with only a pretrained encoder,
such as \cite{resnet}.
Both cases are evaluated in \Cref{sec:ablation}.
\\
\\
\noindent \textbf{Fine-tuning strategies.}
The pretrained $f(\cdot;\theta_{nat})$
can be fine-tuned with different strategies,
ranging from updating all the weights, $\theta_{nat}$, 
to fine-tuning only a subset of layers while keeping the rest frozen.
We adopt Low-Rank Adaptation (LoRA) \cite{lora} as the default fine-tuning strategy for \proposed,
but also investigate alternative options in \Cref{sec:ablation}
to provide a comprehensive comparison.

\subsection{Iterative Co-Training}
\label{sec:training}
Both models, $f(\cdot;\theta_{nat})$ and $g(\cdot;\theta_{med})$,
are then trained on both the labeled subset, $\mathcal{I}_L$ and $\mathcal{M}_L$, as well as the unlabeled subset, $\mathcal{I}_U$,
as shown in \cref{fig:pipeline}.
\\
\\
\noindent \textbf{\emph{Odd}-number epochs.} 
2D slices, $\{{\mathbf{S}^d_{i}}\}_{i=1, d=1}^{n,D}$, 
are extracted along the depth dimension $D$ from 
$\mathcal{I}_U = \{{\mathbf{I}_i}\}_{i=1}^n$,
where $\mathbf{S}^d_i \in \mathbb{R}^{C_i \times H \times W}$.
These slices are input to $f(\cdot;\theta_{nat})$ to generate the pseudo-masks,
$\{{\mathbf{P}_{i}}\}_{i=1}^{n}, \mathbf{P}_i \in \mathbb{R}^{C_c \times H \times W \times D}$:
\begin{equation}
\label{eq:pseudo_2d}
\mathbf{P}_i = concat\left(f(\mathbf{S}^1_{i};\theta_{nat}),\ f(\mathbf{S}^2_{i};\theta_{nat}),\ ...,\ f(\mathbf{S}^D_{i};\theta_{nat})\right),
\end{equation}
where $concat(\cdot)$ concatenates the 2D predicted masks across the depth dimension $D$. 
Together with the masks predicted by $g(\cdot;\theta_{med})$:
\begin{equation}
\label{eq:prediction_3d}
[\hat{\mathbf{M}}_1,\ \hat{\mathbf{M}}_2,...,\ \hat{\mathbf{M}}_n] = [g(\mathbf{I}_1;\theta_{med}),\ g(\mathbf{I}_2;\theta_{med}),\ ...,\ g(\mathbf{I}_n;\theta_{med})],
\end{equation}
the unlabeled loss, $\mathcal{L}_{u}$, can be computed as:
\begin{equation}
\label{eq:loss_unlabel}
\mathcal{L}_u = w_{kl}\cdot\mathcal{L}_{kl}\left(\hat{\mathbf{M}},\mathbf{P}\right) + w_{dice}\cdot\mathcal{L}_{dice}\left(\hat{\mathbf{M}},\mathbf{P}\right),
\end{equation}
where $\mathcal{L}_{kl}$ is the Kullback-Leibler divergence loss and $w_{kl}$ is its weight. 

Supervising with only $\mathcal{L}_u$ may result in collapsed solutions,
for example outputting the same prediction regardless of the inputs.
Therefore, 
the labeled loss, $\mathcal{L}_l$ (\cref{eq:loss_label}),
is also computed,
leading to the final co-training loss, $\mathcal{L}_c$:
\begin{equation}
\label{eq:loss_co}
\mathcal{L}_c = \frac{b_l}{b_l+b_u}\cdot\mathcal{L}_l + \frac{b_u}{b_l+b_u}\cdot\mathcal{L}_u,
\end{equation}
where $b_l$ and $b_u$ are the number of labeled and unlabeled data in a batch.

At each back-propagation step, 
a stochastic optimization is computed to minimize $\mathcal{L}_c$,
with respect to $\theta_{med}$ 
to train $g(\cdot;\theta_{med})$:
\begin{equation}
\label{eq:optimize_3d}
\theta_{med} \leftarrow optim(\theta_{med}, \nabla_{\theta_{med}}\mathcal{L}_c, \eta_{med}),
\end{equation}
where $optim(\cdot)$ indicates the optimizer and $\eta_{med}$ is the learning rate for $g(\cdot;\theta_{med})$.
\\
\\
\noindent \textbf{\emph{Even}-number epochs.}
Pseudo-masks, $\{{\mathbf{P}_{i}}\}_{i=1}^{n}$, are generated by $g(\cdot;\theta_{med})$:
\begin{equation}
\label{eq:pseudo_3d}
[\mathbf{P}_1,\ \mathbf{P}_2,...,\ \mathbf{P}_n] = [g(\mathbf{I}_1;\theta_{med}),\ g(\mathbf{I}_2;\theta_{med}),\ ...,\ g(\mathbf{I}_n;\theta_{med})]
\end{equation}
Meanwhile, $\{{\hat{\mathbf{M}}_i}\}_{i=1}^{n}$ are output by $f(\cdot;\theta_{nat})$ from the input 2D slices, $\{{\mathbf{S}^d_{i}}\}_{i=1, d=1}^{n,D}$:
\begin{equation}
\label{eq:prediction_2d}
\hat{\mathbf{M}}_i = concat\left(f(\mathbf{S}^1_{i};\theta_{nat}),\ f(\mathbf{S}^2_{i};\theta_{nat}),\ ...,\ f(\mathbf{S}^D_{i};\theta_{nat})\right)
\end{equation}
By minimizing the co-training loss, $\mathcal{L}_c$ (\cref{eq:loss_co}),
with respect to $\theta_{nat}$ (or a subset of $\theta_{nat}$ depending on the fine-tuning strategies described in \Cref{sec:finetuning}),
the training of $f(\cdot;\theta_{nat})$ can be summarized as:
\begin{equation}
\label{eq:optimize_2d}
\theta_{nat} \leftarrow optim(\theta_{nat}, \nabla_{\theta_{nat}}\mathcal{L}_c, \eta_{nat})
\end{equation}
%
Through iterating between \cref{eq:optimize_3d,eq:optimize_2d},
the two models, 
$f(\cdot;\theta_{nat})$ and $g(\cdot;\theta_{med})$,
improve their performance by continuously learning from each other.

\subsection{Learning Rate Guided Sampling (LRG-sampling)}
\label{sec:lrg}
Due to the significant imbalance between the size of the labeled set $\mathcal{I}_L$ and that of the unlabeled set $\mathcal{I}_U$,
uniform data sampling from $\mathcal{I}$ during the co-training may undermine the training stability and final performance.
Even with strategies like over-sampling,
a fixed sampling approach may not be optimal.


Ideally,
the proportion of labeled and unlabeled images within a batch
should adapt to the current state of the models.
At early stages when the models' predictions are still unstable and inaccurate,
relying heavily on their generated pseudo-masks may be suboptimal.
As training progresses and the predictions become more stable and accurate,
unlabeled images in a batch, used as pseudo-masks, should increase accordingly
to maximize the utilization of unlabeled data.

This pattern aligns with learning rate decay,
a widely adopted technique for training deep models \cite{bengio2012practical}.
Therefore, we propose learning rate guided (LRG)-sampling,
which adaptively adjusts $b_l$ and $b_u$ (\cref{eq:loss_co}), 
according to the current learning rate, $\eta_{current}$, at each epoch.
This can be formulated as:
\begin{equation}
\label{eq:bu}
b_u = \left\lfloor\frac{\eta_{initial}-\eta_{current}}{\eta_{initial}-\eta_{final}}\cdot B\right\rfloor
\end{equation}
\begin{equation}
\label{eq:bu}
b_l = B-b_u
\end{equation}
where $\lfloor \, \rfloor$ denotes the floor function, 
$B$ is the batch size, and 
$\eta_{initial}$ and $\eta_{final}$ are the initial and final learning rates of the schedule.


\section{Experiments and Results}
\label{sec:experiments}

\subsection{Experimental Settings}
\label{sec:setting}
We conducted a comprehensive evaluation of \proposed,
comparing it with 13 state-of-the-art methods under various settings
(\ie training with different numbers of labeled data) and 
investigated the effect of different components in \proposed through ablation studies.
\\

\noindent \textbf{Datasets.}
We benchmarked on the left atrial (LA) cavity dataset \cite{la} and the Pancreas-CT dataset \cite{pancreas}.
The \textbf{LA dataset} consists of 3D gadolinium-enhanced cardiac MR images,
with ground-truth masks of the LA cavity. 
The resolution is $0.625 \times 0.625 \times 0.625 mm^3$.
We normalized the voxel values of each image to the range $[0, 1]$ and then standardized them.
There are 100 images in the training set and 56 in the testing set.
The \textbf{Pancreas-CT dataset} contains 82 (62 training and 20 testing) abdominal contrast enhanced 3D CT images,
with ground-truth masks of the pancreas.
We re-sampled the images to a unified resolution of $0.85 \times 0.85 \times 0.75 mm^3$.
The voxel values were clipped to the range $[-175,250]$ Hounsfield Units 
and then normalized to $[0, 1]$.
\\

\noindent \textbf{Implementation details.}
\proposed used SegFormer-B2 \cite{segformer},
pretrained on Imagenet-1K \cite{imagenet} and ADE20K \cite{ade20k},
as the 2D model $f(\cdot;\theta_{nat})$,
and a randomly initialized 3D UNet \cite{unet} as the 3D segmentation network $g(\cdot;\theta_{med})$.
Other 2D and 3D models, 
including UNet with pretrained ResNet-50 encoder \cite{resnet,unet} and SwinUNETR \cite{swinunetr},
were also evaluated in ablation studies.
The hyperparameters were set to: 
$w_{ce}=w_{kl}=w_{dice}=1$ and $B=5$.
During training,
we applied a range of data augmentations,
including random resizing with a scale of $0.9-1.1$, 
gamma contrast adjustment with $\gamma\in[0.8,1.2]$
and random cropping to $160 \times 160 \times 64$ with a $50\%$ chance of foreground inclusion.
Pseudo-masks predicted by the 3D model are used to determine the foreground regions for unlabeled images.
The two models were first trained and fine-tuned for 500 epochs (with 50 warm-up epochs) using the labeled images
and then co-trained for another 3500 epochs,
using a weight decay cosine schedule with $\eta_{initial}=10^{-3}$ for $f(\cdot;\theta_{nat})$ and
$\eta_{initial}=10^{-4}$ for $g(\cdot;\theta_{med})$
and $\eta_{final}=0$.
AdamW \cite{adamw} was used as the optimizer.
\\

\noindent \textbf{Inference.}
For each volume, we sampled patches of size 
$160 \times 160 \times 64$, matching the dimensions used during training, 
to ensure full coverage of the volume. 
Adjacent patches were sampled with a 50\% overlap, 
and the predictions within the overlapping regions were averaged to obtain the final output. 
No data augmentation was applied during inference.
\\

\noindent \textbf{Evaluation metrics.}
Four metrics, namely Dice, Jaccard, 95$\%$ Hausdorff Distance ($\text{HD}_{95}$) and Average Surface Distance (ASD), were used for evaluation.
For Dice and Jaccard, higher values indicate better performance,
whereas lower values are desirable for $\text{HD}_{95}$ and ASD.


\begin{table}[!t]
    \fontsize{8}{10}\selectfont
    \parbox[t][][]{.50\linewidth}{
        \caption{LA results (8 labels)} 
        \centering
        \begin{tabular}{ccccc}
\hline
\multicolumn{1}{c|}{\multirow{2}{*}{Method}}  & \multicolumn{4}{c}{Metrics}                                                                 \\ \cline{2-5} 
\multicolumn{1}{c|}{} &
  \multicolumn{1}{c|}{\begin{tabular}[c]{@{}c@{}}Dice\\ (\%)$\uparrow$\end{tabular}} &
  \multicolumn{1}{c|}{\begin{tabular}[c]{@{}c@{}}Jaccard\\ (\%)$\uparrow$\end{tabular}} &
  \multicolumn{1}{c|}{\begin{tabular}[c]{@{}c@{}}$\text{HD}_{95}$\\ (vox)$\downarrow$\end{tabular}} &
  \begin{tabular}[c]{@{}c@{}}ASD\\ (vox)$\downarrow$\end{tabular} \\ \hline
\multicolumn{1}{c|}{3D UNet}                  & \multicolumn{1}{c|}{88.26} & \multicolumn{1}{c|}{79.89} & \multicolumn{1}{c|}{11.19} & 3.29 \\
\multicolumn{1}{c|}{*SegFormer}               & \multicolumn{1}{c|}{83.05} & \multicolumn{1}{c|}{73.25} & \multicolumn{1}{c|}{15.79} & 5.40 \\
\multicolumn{1}{c|}{**SegFormer}              & \multicolumn{1}{c|}{88.61} & \multicolumn{1}{c|}{79.87} & \multicolumn{1}{c|}{10.49} & 3.35 \\ \hline
\multicolumn{1}{c|}{SASSNet\cite{sassnet}}    & \multicolumn{1}{c|}{87.32} & \multicolumn{1}{c|}{77.72} & \multicolumn{1}{c|}{9.62}  & 2.55 \\
\multicolumn{1}{c|}{MC-Net\cite{mcnet}}       & \multicolumn{1}{c|}{87.71} & \multicolumn{1}{c|}{78.31} & \multicolumn{1}{c|}{9.36}  & 2.18 \\
\multicolumn{1}{c|}{SS-Net\cite{ssnet}}       & \multicolumn{1}{c|}{88.55} & \multicolumn{1}{c|}{79.62} & \multicolumn{1}{c|}{7.49}  & 1.9  \\
\multicolumn{1}{c|}{MC-Net+\cite{mcnetp}}     & \multicolumn{1}{c|}{88.96} & \multicolumn{1}{c|}{80.25} & \multicolumn{1}{c|}{7.93}  & 1.86 \\
\multicolumn{1}{c|}{CAML\cite{caml}}          & \multicolumn{1}{c|}{89.62} & \multicolumn{1}{c|}{81.28} & \multicolumn{1}{c|}{8.76}  & 2.02 \\
\multicolumn{1}{c|}{DK-UXNet\cite{dkuxnet}}   & \multicolumn{1}{c|}{90.41} & \multicolumn{1}{c|}{82.69} & \multicolumn{1}{c|}{7.32}  & 1.71 \\
\multicolumn{1}{c|}{UA-MT\cite{uamt}}         & \multicolumn{1}{c|}{84.25} & \multicolumn{1}{c|}{73.48} & \multicolumn{1}{c|}{13.84} & 3.36 \\
\multicolumn{1}{c|}{LG-ER-MT\cite{lgermt}}    & \multicolumn{1}{c|}{85.54} & \multicolumn{1}{c|}{75.12} & \multicolumn{1}{c|}{13.29} & 3.77 \\
\multicolumn{1}{c|}{DUWM\cite{duwm}}          & \multicolumn{1}{c|}{85.91} & \multicolumn{1}{c|}{75.75} & \multicolumn{1}{c|}{12.67} & 3.31 \\
\multicolumn{1}{c|}{AD-MT\cite{admt}}         & \multicolumn{1}{c|}{90.55} & \multicolumn{1}{c|}{82.79} & \multicolumn{1}{c|}{5.81}  & 1.7  \\
\multicolumn{1}{c|}{BCP\cite{bcp}}            & \multicolumn{1}{c|}{89.62} & \multicolumn{1}{c|}{81.31} & \multicolumn{1}{c|}{6.81}  & 1.76 \\
\multicolumn{1}{c|}{Co-BioNet\cite{cobionet}} & \multicolumn{1}{c|}{89.2}  & \multicolumn{1}{c|}{80.68} & \multicolumn{1}{c|}{6.44}  & 1.9  \\
\multicolumn{1}{c|}{GraphCL\cite{graphcl}}    & \multicolumn{1}{c|}{90.24} & \multicolumn{1}{c|}{82.31} & \multicolumn{1}{c|}{6.42}  & 1.71 \\ \hline
\multicolumn{1}{c|}{\begin{tabular}[c]{@{}c@{}}\proposed\\ (Ours)\end{tabular}} &
  \multicolumn{1}{c|}{\textbf{91.56}} &
  \multicolumn{1}{c|}{\textbf{84.47}} &
  \multicolumn{1}{c|}{\textbf{4.59}} &
  \textbf{1.40} \\ \hline
\multicolumn{5}{l}{\begin{tabular}[c]{@{}l@{}}\textbf{bold} indicates top performance\\ $\uparrow$ means higher values being more accurate\\ * SegFormer trained from scratch\\ ** SegFormer pretrained on ADE20K \cite{ade20k}\end{tabular}}
\end{tabular}
        \label{tab:la8}
    }
    \hfill
    \parbox[t][][]{.50\linewidth}{
        \caption{LA results (4 labels)}
        \centering
        \begin{tabular}{c|cccc}
\hline
\multirow{2}{*}{Method} & \multicolumn{4}{c}{Metrics}                                                                              \\ \cline{2-5} 
                        & \multicolumn{1}{c|}{Dice}  & \multicolumn{1}{c|}{Jaccard} & \multicolumn{1}{c|}{$\text{HD}_{95}$} & ASD  \\ \hline
3D UNet                 & \multicolumn{1}{c|}{82.01} & \multicolumn{1}{c|}{72.42}   & \multicolumn{1}{c|}{29.95}            & 9.91 \\
*SegFormer              & \multicolumn{1}{c|}{71.01} & \multicolumn{1}{c|}{58.4}    & \multicolumn{1}{c|}{14.34}            & 7.07 \\
**SegFormer             & \multicolumn{1}{c|}{84.23} & \multicolumn{1}{c|}{74.49}   & \multicolumn{1}{c|}{8.24}             & 3.57 \\ \hline
SS-Net\cite{ssnet}      & \multicolumn{1}{c|}{86.33} & \multicolumn{1}{c|}{76.15}   & \multicolumn{1}{c|}{9.97}             & 2.31 \\
CAML\cite{caml}         & \multicolumn{1}{c|}{87.34} & \multicolumn{1}{c|}{77.65}   & \multicolumn{1}{c|}{9.76}             & 2.49 \\
DK-UXNet\cite{dkuxnet} & \multicolumn{1}{c|}{85.96}          & \multicolumn{1}{c|}{75.91}          & \multicolumn{1}{c|}{11.72}         & 2.64          \\
AD-MT\cite{admt}        & \multicolumn{1}{c|}{89.63} & \multicolumn{1}{c|}{81.28}   & \multicolumn{1}{c|}{6.56}             & 1.85 \\
BCP\cite{bcp}           & \multicolumn{1}{c|}{88.02} & \multicolumn{1}{c|}{78.72}   & \multicolumn{1}{c|}{7.9}              & 2.15 \\
GraphCL\cite{graphcl}   & \multicolumn{1}{c|}{88.8}  & \multicolumn{1}{c|}{80}      & \multicolumn{1}{c|}{7.16}             & 2.1  \\ \hline
\proposed              & \multicolumn{1}{c|}{\textbf{90.47}} & \multicolumn{1}{c|}{\textbf{82.66}} & \multicolumn{1}{c|}{\textbf{4.72}} & \textbf{1.59} \\ \hline
\end{tabular}
         \label{tab:la4}
         \medskip

        \caption{Pancreas-CT results (6 labels)}
        \centering
        \begin{tabular}{c|cccc}
\hline
\multirow{2}{*}{Method} & \multicolumn{4}{c}{Metrics}                                                                              \\ \cline{2-5} 
                        & \multicolumn{1}{c|}{Dice}  & \multicolumn{1}{c|}{Jaccard} & \multicolumn{1}{c|}{$\text{HD}_{95}$} & ASD  \\ \hline
3D UNet                 & \multicolumn{1}{c|}{59.53} & \multicolumn{1}{c|}{44.17}   & \multicolumn{1}{c|}{18.52}            & 1.79 \\
*SegFormer              & \multicolumn{1}{c|}{43.76} & \multicolumn{1}{c|}{28.32}   & \multicolumn{1}{c|}{19.67}            & 6.86 \\
**SegFormer             & \multicolumn{1}{c|}{68.57} & \multicolumn{1}{c|}{53.34}   & \multicolumn{1}{c|}{16.39}            & 4.77 \\ \hline
MC-Net\cite{mcnet}      & \multicolumn{1}{c|}{68.94} & \multicolumn{1}{c|}{54.74}   & \multicolumn{1}{c|}{16.28}            & 3.16 \\
MC-Net+\cite{mcnetp}    & \multicolumn{1}{c|}{74.01} & \multicolumn{1}{c|}{60.02}   & \multicolumn{1}{c|}{12.59}            & 3.34 \\
AD-MT\cite{admt}        & \multicolumn{1}{c|}{80.21} & \multicolumn{1}{c|}{67.51}   & \multicolumn{1}{c|}{7.18}             & 1.66 \\
Co-BioNet\cite{cobionet} & \multicolumn{1}{c|}{77.89}          & \multicolumn{1}{c|}{64.79}          & \multicolumn{1}{c|}{8.81}          & \textbf{1.39} \\ \hline
\proposed                & \multicolumn{1}{c|}{\textbf{81.67}} & \multicolumn{1}{c|}{\textbf{69.53}} & \multicolumn{1}{c|}{\textbf{6.56}} & 1.67          \\ \hline
\end{tabular}
         \label{tab:ct6}
         
         }
\end{table}

\begin{figure*}[h]
    \centering 
    \includegraphics[width=1\textwidth]{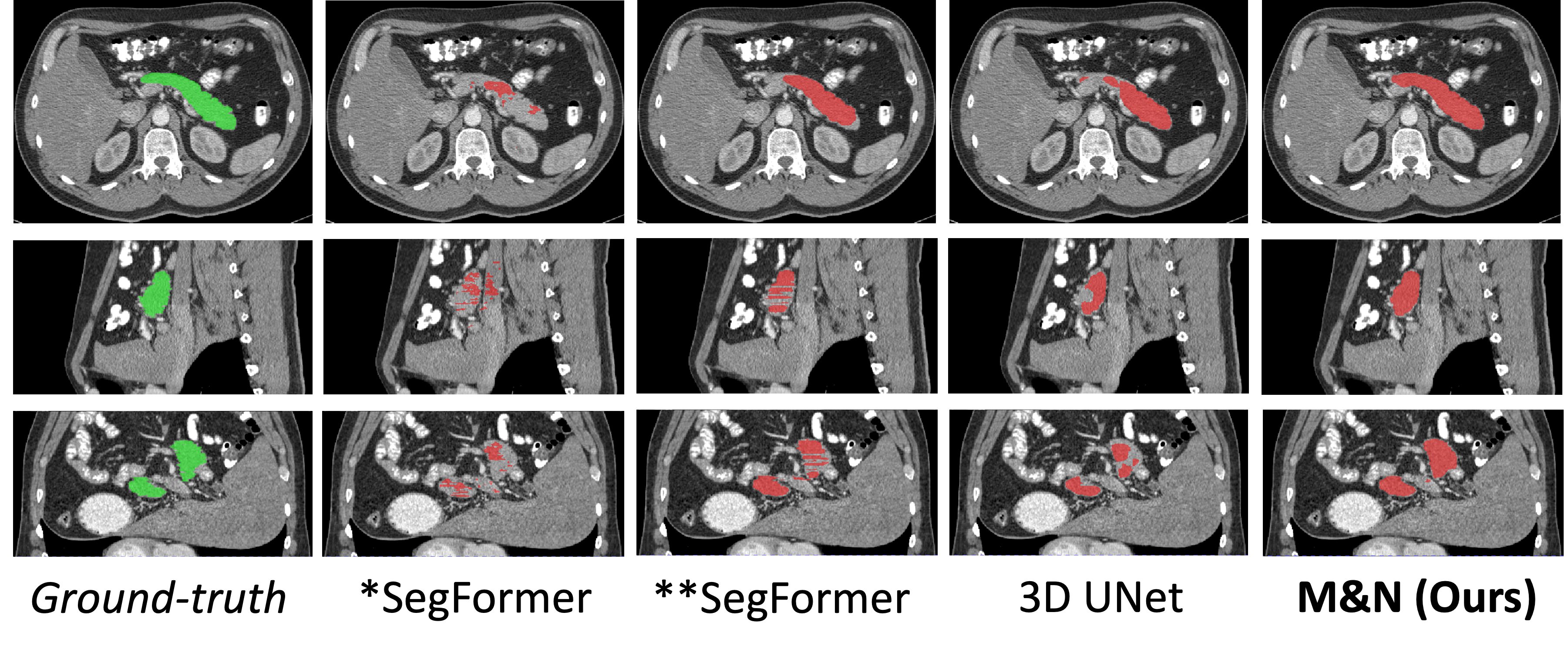}
    \caption{Qualitative results of Pancreas-CT dataset \cite{pancreas}.
             All approaches are trained from the same number (\ie 6) of labeled images.*SegFormer is trained from scratch and **SegFormer is first pretrained on ADE20K \cite{ade20k}.
    }
\label{fig:qualitative}
\end{figure*}

\subsection{Baseline Comparisons}
\label{sec:baseline}
We compared \proposed with 13 state-of-the-art approaches on the LA dataset, trained with 4 and 8 labeled images, 
and the Pancreas-CT dataset, trained with 6 labeled images.

As shown in \cref{tab:la8,tab:la4,tab:ct6},
our proposed \proposed consistently outperformed all existing approaches across different datasets and settings (\ie results in \textbf{bold}).
Testing on different modalities (\ie MRI and CT) and target structures (\ie LA and pancreas) using different numbers of labeled training data,
our results demonstrated the robustness and generalizability of \proposed.

The results in the first three rows of the tables, 
particularly \cref{tab:ct6},
and \cref{fig:qualitative},
which present the more challenging CT pancreas segmentation task,
validated the motivation behind our work.
Specifically, pretraining on 2D natural images significantly helps medical segmentation (\ie *SegFormer vs. **SegFormer),
whereas 3D UNet trained with only small amount of labeled data struggled to achieve satisfactory results. 
However, it still outperformed a 2D network trained from scratch (\ie 3D UNet vs. *SegFormer),
justifying \proposed's choice of distilling knowledge from a pretrained 2D model to a 3D model.

In comparison to the best existing method, AD-MT \cite{admt},
we used 3D UNet while AD-MT used VNet \cite{vnet} as the segmentation model.
These two models share very similar architectures,
further verifying that the better performance of \proposed is primarily attributed to the differences in the learning framework design rather than the network architecture.

We focused on the low-labeled data regime,
where fewer than 10 labeled images are available (\ie 4, 6, and 8).
This threshold was deliberately chosen based on our previous experience collaborating with clinicians,
whose incentive to annotate data significantly drops when the required number increases by an order of magnitude (\eg 9 to 10 or 90 to 100).
This phenomenon is consistent with psychological and marketing concepts, 
such as categorical perception of numbers and the left-digit effect \cite{anderson2003effects}.
By limiting our evaluation to fewer than 10 labeled images,
our experiment setting aimed to reflect realistic scenarios where annotation resources are scarce.

\subsection{Ablation Studies}
\label{sec:ablation}
We investigated the effect of different components in \proposed on the LA dataset,
trained with 8 labeled images, 
with the results presented in \Cref{tab:ablation}.
\\

\noindent \textbf{Model-agnostic analysis.}
While the default pretrained SegFormer has an encoder-decoder architecture, 
we replaced it with a pretrained ResNet-50 \cite{resnet} encoder, 
connected to a randomly initialized decoder through skip connections, 
termed as ResUNet.
As shown in \cref{tab:ablation} (row 1), its performance slightly decreased but still outperformed the best existing method, AD-MT \cite{admt} (\cref{tab:la8}).
Additionally, 
we replaced 3D UNet with SwinUNETR \cite{swinunetr}
and observed a similar slight drop in performance (\cref{tab:ablation} row 2),
but still surpassing AD-MT \cite{admt} on most metrics.
These experiments validated the model-agnostic nature of \proposed,
confirming its ability to adapt to various architectures while maintaining superior performance.
\\

\noindent \textbf{Fine-tuning strategies.}
We compared 3 fine-tuning strategies for the pretrained SegFormer, 
namely LoRA \cite{lora}, decoder fine-tuning and whole network fine-tuning.
While freezing the encoder and updating only the decoder resulted in a performance drop (row 5),
the other two strategies performed comparably (row 6),
with LoRA slightly outperformed on 3 out of 4 metrics.
Since LoRA is also more parameter-efficient,
we adopted it as the default strategy for \proposed.
\\

\noindent \textbf{Training and data sampling.}
We ablated iterative co-training by using fixed pseudo-masks training,
where the pre-trained 2D model was fine-tuned with labeled images (\Cref{sec:finetuning})
and then used to generate pseudo-masks for training the 3D model without further updates.
This led to a performance decrease (row 3).
Additionally, replacing LRG-sampling with uniform training data sampling resulted in an even more significant drop in performance (row 4).
These experiments validated the effectiveness of our proposed components in \proposed.

\begin{table}[t]
\fontsize{8}{10}\selectfont
\caption{Ablation studies of \proposed on LA datasets (8 labels).} 
\label{tab:ablation}
\begin{tabular}{ccccccccc}
\hline
\multicolumn{1}{c|}{\multirow{2}{*}{\begin{tabular}[c]{@{}c@{}}Pretrained\\ 2D model\end{tabular}}} &
  \multicolumn{1}{l|}{\multirow{2}{*}{\begin{tabular}[c]{@{}l@{}}Fine-tuning \\ strategy*\end{tabular}}} &
  \multicolumn{1}{l|}{\multirow{2}{*}{3D network}} &
  \multicolumn{1}{l|}{\multirow{2}{*}{\begin{tabular}[c]{@{}l@{}}Iterative\\ co-training\end{tabular}}} &
  \multicolumn{1}{l|}{\multirow{2}{*}{\begin{tabular}[c]{@{}l@{}}LRG-\\ sampling\end{tabular}}} &
  \multicolumn{4}{c}{Metrics} \\ \cline{6-9} 
\multicolumn{1}{c|}{} &
  \multicolumn{1}{l|}{} &
  \multicolumn{1}{l|}{} &
  \multicolumn{1}{l|}{} &
  \multicolumn{1}{l|}{} &
  \multicolumn{1}{c|}{\begin{tabular}[c]{@{}c@{}}Dice\\ (\%)$\uparrow$\end{tabular}} &
  \multicolumn{1}{c|}{\begin{tabular}[c]{@{}c@{}}Jaccard\\ (\%)$\uparrow$\end{tabular}} &
  \multicolumn{1}{c|}{\begin{tabular}[c]{@{}c@{}}$\text{HD}_{95}$\\ (vox)$\downarrow$\end{tabular}} &
  \begin{tabular}[c]{@{}c@{}}ASD\\ (vox)$\downarrow$\end{tabular} \\ \hline
ResUNet** &
  Whole &
  3D UNet &
  \cmark &
  \multicolumn{1}{c|}{\cmark} &
  \multicolumn{1}{c|}{91.11} &
  \multicolumn{1}{c|}{83.74} &
  \multicolumn{1}{c|}{5.38} &
  1.41 \\
SegFormer &
  LoRA &
  SwinUNETR &
  \cmark &
  \multicolumn{1}{c|}{\cmark} &
  \multicolumn{1}{c|}{90.67} &
  \multicolumn{1}{c|}{83.03} &
  \multicolumn{1}{c|}{5.24} &
  2.16 \\
SegFormer &
  LoRA &
  3D UNet &
  \xmark &
  \multicolumn{1}{c|}{\cmark} &
  \multicolumn{1}{c|}{89.4} &
  \multicolumn{1}{c|}{81.32} &
  \multicolumn{1}{c|}{6.06} &
  1.74 \\
SegFormer &
  LoRA &
  3D UNet &
  \cmark &
  \multicolumn{1}{c|}{\xmark} &
  \multicolumn{1}{c|}{87.14} &
  \multicolumn{1}{c|}{77.35} &
  \multicolumn{1}{c|}{8.43} &
  2.36 \\
SegFormer &
  Decoder &
  3D UNet &
  \cmark &
  \multicolumn{1}{c|}{\cmark} &
  \multicolumn{1}{c|}{89.49} &
  \multicolumn{1}{c|}{81.28} &
  \multicolumn{1}{c|}{6.44} &
  1.75 \\
SegFormer &
  Whole &
  3D UNet &
  \cmark &
  \multicolumn{1}{c|}{\cmark} &
  \multicolumn{1}{c|}{91.39} &
  \multicolumn{1}{c|}{84.20} &
  \multicolumn{1}{c|}{5.02} &
  \textbf{1.38} \\ \hline
SegFormer &
  LoRA &
  3D UNet &
  \cmark &
  \multicolumn{1}{c|}{\cmark} &
  \multicolumn{1}{c|}{\textbf{91.56}} &
  \multicolumn{1}{c|}{\textbf{84.47}} &
  \multicolumn{1}{c|}{\textbf{4.59}} &
  1.40 \\ \hline
\multicolumn{9}{l}{\begin{tabular}[c]{@{}l@{}}* Finetuning strategy refers to which part of the pretrained 2D model we update. \\ ** UNet with ResNet-50 encoder pretrained on Imagenet-1K \cite{imagenet}\end{tabular}}
\end{tabular}
\end{table}

\section{Conclusion}
In summary, 
we present \proposed,
a model-agnostic framework for transferring the knowledge from
general vision models pretrained on 2D natural images 
to enhance semi-supervised 3D medical image segmentation.
By iteratively co-training the 2D and 3D models and adaptively adjusting the proportion of labeled and unlabeled images within a batch throughout training,
\proposed achieves state-of-the-art performance on various publicly available datasets under different limited labeled data settings, 
outperforming 13 existing methods.
As future work,
we plan to extend \proposed to other tasks in medical image analysis,
for example image registration.
Our ultimate goal is to utilize the abundant cross-domain knowledge to facilitate the development in medical image analysis.

\subsubsection{Acknowledgments}
PH. Yeung is funded by the Presidential Postdoctoral Fellowship from Nanyang Technological University. 
We thank Dr Madeleine Wyburd and Mr Valentin Bacher for their valuable suggestions and comments about the work.

\bibliographystyle{splncs04}
\bibliography{ref}
\end{document}